\documentclass[letterpaper, 10 pt, journal, twoside]{IEEEtran}

\IEEEoverridecommandlockouts        

\usepackage{graphics} 
\usepackage{epsfig} 
\usepackage{amsmath, amssymb} 

\usepackage{booktabs} 
\usepackage{hyperref}
\usepackage{multirow} 
\usepackage{subcaption}
\usepackage{algorithm} 
\usepackage{algpseudocode}

\algrenewcommand\algorithmicrequire{\textbf{Input:}}
\algrenewcommand\algorithmicensure{\textbf{Output:}}

\def\R{\mathbb{R}}

\DeclareMathOperator*{\argmin}{arg\,min}

\title{Manipulation of Granular Materials by Learning Particle Interactions}

\author{Neea Tuomainen*, David Blanco-Mulero*, Ville Kyrki
\thanks{Manuscript accepted for publication at IEEE Robotics and Automation Letters.}%
\thanks{
This work was financially supported by the Academy of Finland grant numbers 317020 and 328399. (\textit{Corresponding author: David Blanco-Mulero.})}%
\thanks{* Neea Tuomainen and David Blanco-Mulero are co-first authors and contributed to this work equally.}%
\thanks{The authors are with the Department of Electrical Engineering and Automation (EEA), Aalto University, 02150 Espoo, Finland.
(e-mail: { neea.tuomainen@aalto.fi, david.blancomulero@aalto.fi, ville.kyrki@aalto.fi})}
}

\begin{document}

\maketitle
\thispagestyle{empty}
\pagestyle{empty}

\begin{abstract}
Manipulation of granular materials such as sand or rice remains an unsolved problem due to challenges such as the difficulty of defining their configuration or modeling the materials and their particles interactions.
Current approaches tend to simplify the material dynamics and omit the interactions between the particles.
In this paper, we propose to use a graph-based representation to model the interaction dynamics of the material and rigid bodies manipulating it.
This allows the planning of manipulation trajectories to reach a desired configuration of the material.
We use a graph neural network (GNN) to model the particle interactions via message-passing.
To plan manipulation trajectories, we propose to minimise the Wasserstein distance between a predicted distribution of granular particles and their desired configuration.
We demonstrate that the proposed method is able to pour granular materials into the desired configuration both in simulated and real scenarios.
\end{abstract}


\section{Introduction}
The concept of granular materials encompasses materials such as ground coffee, uncooked rice, and sand to name a few.
These materials are difficult to manipulate due to their particle nature \cite{Yin_2017_granular_material_mechanics}, as the particles of the material interact with each other potentially grouping into individual shapes. 
Recently, several attempts have been made to solve various tasks involving granular materials ranging from transporting the material \cite{Schenck2017_manipulation_granular}, shaping \cite{Cherubini2018_vision_manipulation_plastic, Cherubini2020_modelfree_shaping}, to common daily tasks such as gathering, spreading or flipping \cite{Zhang2020_amorphous_mat}. 
A common approach is to use visual feedback to learn to manipulate the material.
However, there is a common caveat on these approaches, the interactions of the material particles are not captured. 
Thus, these methods are not able to plan manipulation of granular materials into a desired configuration when there are complex interactions including e.g. dropping the material and the material  interacting with rigid bodies. 
This is because learning the interactions from visual-feedback is not feasible as the material is not fully observable.

Instead of using models learned from direct observations, simulations of the physical interaction can be used.
However, high-fidelity simulators are considerably expensive computationally.
Instead, we can learn a computational surrogate model from these simulators to speed up the manipulation planning.
Recently, graph neural networks (GNNs) have been proposed for modeling interactions between particles \cite{Battaglia2016_interaction, Mrowca2018_hrn_flexible_representation, Battaglia2018_relational, Li2019_learning_particle_dynamics}.
Some of these works have explored modeling the dynamics of complex materials such as granular \cite{Sanchez-Gonzalez2020_learning_to_simulate} or deformable ones \cite{pfaff_2021_mesh-basedlearning}, \cite{lin_2021_vcd}.
The GNN based approaches have shown good accuracy and generalisation capabilities, and thus present a promising surrogate model for granular manipulation planning.

\begin{figure}
    \centering
    \vspace{0.10cm}
    \includegraphics[width=0.9\columnwidth]{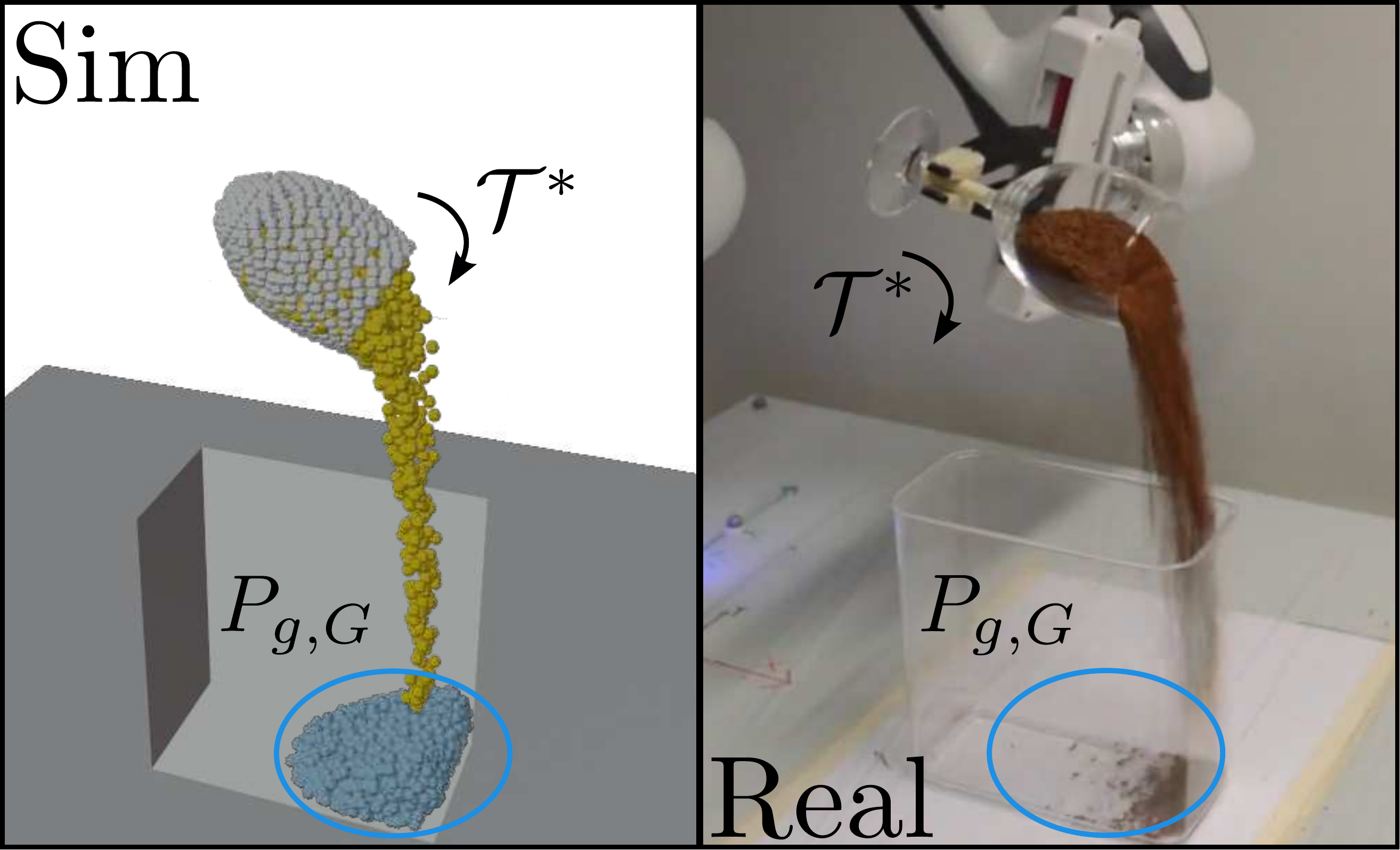}
    \caption{Granular manipulation using the GNN (\textbf{left}) and physical system (\textbf{right}). The granular material is manipulated using a rigid-body (cup) following the optimal trajectory $\mathcal{T}^*$ to pour the material in the particles target distribution $P_{g,G}$.}
    \label{fig:intro_figure}
\end{figure}

In this work, we address the problem of manipulating granular materials into a desired configuration by interacting with a rigid-object (see Fig.~\ref{fig:intro_figure}).
In order to capture the interactions of the granular material with the rigid-body, we take on a model-based approach and learn a surrogate model from simulated data based on the idea of Graph Networks-based Simulators (GNS) \cite{Sanchez-Gonzalez2020_learning_to_simulate}.
We propose to solve the problem of manipulating the material into a desired configuration as a trajectory planning problem, where we use the GNS to rollout the dynamics of the system.
The trajectory of the rigid-body is planned by minimising the Wasserstein distance between the simulated final distribution of granular particles and its target.

We evaluate two different aspects of the method.
First, we perform an ablation study of the graph attributes and GNN architecture, and provide evidence that the chosen attributes and architecture are able to capture the rigid-body and granular material interactions.
Secondly, we evaluate the proposed trajectory planning method over different target distributions of the granular particles, both using the GNN rollout and a real robot set-up (Fig.~\ref{fig:intro_figure}).
As a test-case task we perform pouring of a granular material into a desired configuration.
Our results demonstrate that the proposed method effectively pours the material into different desired distributions.
Furthermore, we demonstrate that the optimal trajectories in the real robot approximate the expected result using the GNN dynamics model.

The main contributions of this paper are:
\begin{enumerate}
    \item A GNN based approach for modeling the interaction of granular materials and rigid bodies, and evaluation of the model parameters required to effectively predict the interactions.
    \item A solution for trajectory planning based on the Wasserstein distance between the distribution of the manipulated granular material and its target distribution.
    \item A demonstration of the framework for pouring granular materials into a desired configuration both in simulated and real scenarios. 
\end{enumerate}

\section{Related work}
\label{sec:background}
\subsection{Granular manipulation}

The manipulation of granular materials has been studied in different tasks such as robotic excavation \cite{Singh1995_robotic_excavation}, scooping  \cite{Sarata2004_trajectory_scooping} and pouring \cite{Clarke2018_audiofeedback_scooping}.
Recent works have focused on learning to manipulate granular materials using feedback-based methods where data is collected in the real-world \cite{Schenck2017_manipulation_granular, Cherubini2018_vision_manipulation_plastic, Cherubini2020_modelfree_shaping, Clarke2018_audiofeedback_scooping, Takahashi2021_grasping_granular}.
In \cite{Schenck2017_manipulation_granular}, a ConvNet model is learned to choose the required actions for transporting beans to a goal shape.
Similarly, a learning-based approach was proposed by \cite{Takahashi2021_grasping_granular} to grasp granular foods.
These methods learn a surrogate model from height or density maps, which does not capture the interaction of the granular particles.
Furthermore, collecting data in the real-world is prohibitive and the feedback from the real data does not capture the inherent dynamics of the material.
More recently, \cite{Zhang2020_amorphous_mat} proposed to learn a Reinforcement Learning policy to manipulate in simulation amorphous materials.
However, the policy takes as input the density and height map, facing the same limitation of previous approaches.
In contrast, \cite{matl_2020_infering_granular_properties} proposed a framework to infer the properties of granular materials using real-world data for fine tuning a simulator to perform robotic tasks.

In the present work, we propose to solve granular manipulation directly in simulation. This alleviates the burden of gathering data in the real-world of previous approaches.
In addition, the proposed GNN approach enables learning the granular material interactions, as opposed to the aforementioned surrogate models. 
Furthermore, our method defines the desired configuration of granular particles as an experimental distribution. This enables planning for target particle configurations potentially more complex than those given by height or density maps.

\subsection{Liquid manipulation}

Similar approaches to those in granular manipulation have been proposed for pouring liquids \cite{Schenck_2017_visual_closed_loop_pouring, Do2018_accurate_pouring, Huang2021_robot_gaining}. 
These either used visual feedback to detect the liquid \cite{Schenck_2017_visual_closed_loop_pouring, Do2018_accurate_pouring} or learned to manipulate from human demonstrations  \cite{Huang2021_robot_gaining}.
While dynamics may be learned on from real examples, using real-world data is limited by data collection, as with granular materials, which can be solved by using simulated dynamics.
Some of the works in liquid manipulation have considered the dynamics of the material either using high-complexity simulations \cite{Pan2016_motion_planning_pouring} or simplified dynamics models \cite{Aribowo2010_input_shaping_control, Reyhanoglu2013_modeling_control_slosh, Pan2016_motion_planning_fluid}. 
Considering the dynamics of the system allows trajectory planning in more complex manipulation tasks.
However, simplified dynamics models might lead to inaccurate trajectories, whereas using high-quality simulations might be costly for complex materials.

\begin{figure*}
    \centering
    \vspace{0.05cm}
    \includegraphics{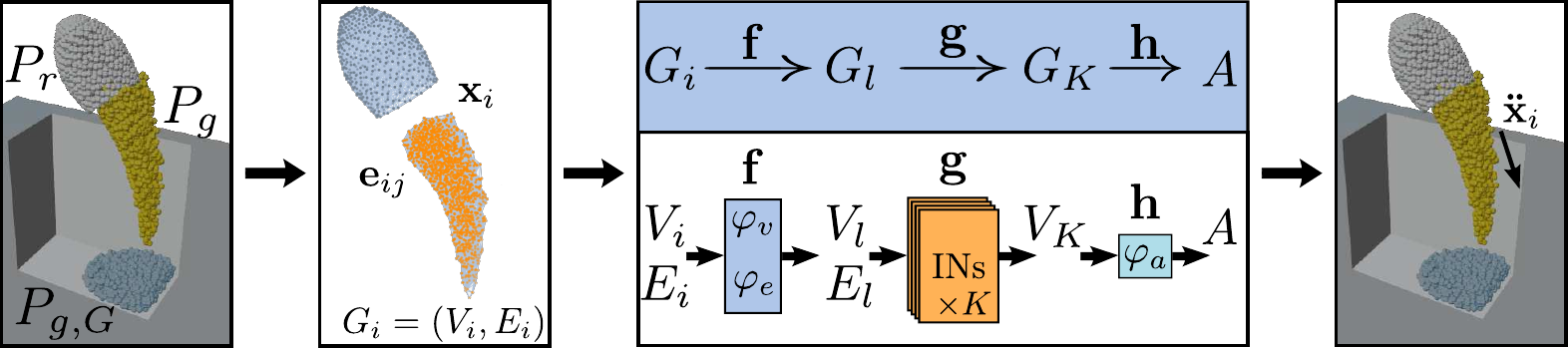}
    \caption{The initial graph $G_i=(V_i, E_i)$ is created from the particles of the granular material $P_g$ and the rigid-body $P_r$. The particles are manipulated to match the goal particle distribution $P_{g,G}$. The graph $G_i$ is mapped through the GNN model encoder $\mathbf{f}$, processor $\mathbf{g}$ and decoder $\mathbf{h}$ to predict the acceleration $\mathbf{\ddot{x}}_i \in A$ of the granular material particles.}
    \label{fig:gnn-parts}
\end{figure*}

\subsection{Learning particle interactions using GNNs}
An alternative is to learn the dynamics of the system from high-fidelity simulators using data-driven models \cite{Li2019_learning_particle_dynamics, Sanchez-Gonzalez2020_learning_to_simulate, pfaff_2021_mesh-basedlearning}. These can learn from simulation techniques and present the benefit of lower computational cost than the simulators they learn from.
Particularly, some GNN approaches have shown outstanding results on learning to simulate complex materials \cite{Sanchez-Gonzalez2020_learning_to_simulate, pfaff_2021_mesh-basedlearning}.
Sanchez-Gonzalez \textit{et al.} \cite{Sanchez-Gonzalez2020_learning_to_simulate} proposed the Graph Networks-based Simulators (GNS) framework, which demonstrated accurate forecasting for different materials such as sand or fluids.
However, their work does not involve modeling the interactions when the materials are manipulated.
Recently, the GNS framework was used to learn the dynamics of manipulated cloths \cite{lin_2021_vcd}, which suggests it is a good candidate for learning to manipulate granular materials.
The closer work to our aim is DPI-Nets \cite{Li2019_learning_particle_dynamics}, as it combines a GNN model for learning material dynamics with model predictive control for controlling a fluid. However, they do not consider the interaction dynamics of pouring the material and the fluid does not have the mechanical properties of granular materials.

\section{Method}
\label{sec:method}
In this paper, we address the problem of manipulation of granular materials that are shaped into a desired configuration by interacting with them using a rigid-object.
We distinguish between two types of materials: the rigid-body that manipulates the granular material, and the granular material itself.
We model both materials as particles, which has proven as a good representation for modelling the interaction between different materials \cite{Mrowca2018_hrn_flexible_representation, Li2019_learning_particle_dynamics,  Sanchez-Gonzalez2020_learning_to_simulate}.
Rigid bodies are modelled as a set of rigid-body particles $\mathbf{p}_r \in P_r$ controllable, where the true position is assumed known.
Granular material particles $\mathbf{p}_g \in P_g$  are affected by gravity, their interaction with each other and with rigid-body particles, and the boundaries of the manipulation scene.
The granular particles cannot be directly controlled, but we can affect their state through their interaction with controlled rigid-body particles.

\subsection{Graph-based representation}
\label{sec:graph_represent}

Our method makes use of a graph representation where each particle $\mathbf{p}_i$ is represented by a node in a graph $G$. The nodes are connected via edges if the particles they represent interact with each other.
We assume an interaction exists when two particles are in contact, that is, their distance is lower than a connectivity radius $R$ (see Section~\ref{sec:exp-gnn-details}).

We define a graph as $G=(V,E)$, where $\mathbf{v}_i \in V$ is a node attribute vector and $\mathbf{e}_{ij} \in E$ represents an edge attribute vector, where $i$ is the index of the sender node and $j$ is the index of the receiver node.
Here, we represent the node set as a matrix $V \in \mathbb{R}^{N \times D}$, where the i-th column represents the node attribute for node $i$, $N$ represents the number of nodes and $D$ the node attributes. Similarly, we represent the edge set as a matrix $E \in \mathbb{R}^{M \times F}$, where $M$ is the number of edges and $F$ the number of edge attributes.
We define the node attribute vector as $\mathbf{v}_i = [ \mathbf{\dot{x}}_{t-C},  \cdots, \mathbf{\dot{x}}_{t-1}, \mathbf{b}_t, m, \mathbf{c}_t ]$, where $\mathbf{\dot{x}}_{t-C},  \cdots, \mathbf{\dot{x}}_{t-1}$ are the $C$ previous velocities of the particle, $\mathbf{b}_t$ represents the relative distance to the manipulation scene boundaries, $m$ denotes whether the particle is a rigid-body or a granular material particle, and $\mathbf{c}_t$ represents the control input applied to the particle.
The control input is defined as the current velocity of the particle $\mathbf{\dot{x}_{t}}$ for rigid-body particles, and as a zero vector for granular material particles.
The edge attribute vector is $ \mathbf{e}_{ij} = [ \mathbf{s}_{ij}, d_{ij}] $, where $\mathbf{s}_{ij}$ represents the relative displacement between the particles $\mathbf{s}_{ij}=\mathbf{p}_i - \mathbf{p}_j$, and $d_{ij}$ represents their relative distance.

\subsection{Graph Neural Network model}
\label{sec:gnn_model}

We learn the dynamics of the granular material particles and their interaction with the rigid-body using a GNN model.
The GNN model is used to predict the acceleration of the granular material particles and their trajectories assuming their initial position as well as the position of the rigid-body particles are known.
Our network architecture is based on the  GNS framework \cite{Sanchez-Gonzalez2020_learning_to_simulate}.
The model is divided into three sequential parts: (a) an encoder $\mathbf{f}: G_{i} \mapsto G_{l}$, (b) a processor $\mathbf{g}: G_{l} \mapsto G_{K}$, and (c) a decoder $\mathbf{h}: G_{K} \mapsto A$.
The encoder maps the input graph $G_{i}$ into a latent graph $G_{l}$.
The processor propagates the information through $K$ latent graphs, outputting the final latent graph $G_{K}$.
Finally, the decoder produces as output the acceleration of each particle represented by a node using the node attributes of the final latent graph $G_{K}$.
Each sequential part is built as follows:

\paragraph{Encoder} The encoder takes an input graph $G_{i} = (V_{i}, E_{i}) $ and encodes the node attributes $V_{i} \in \R^{N \times D_{i}}$ and edge attributes $E_{i} \in \mathbb{R}^{M \times F_{i}}$ into a latent graph $G_{l} =  (V_{l}, E_{l})$, where $V_{l} \in \R^{N \times D_{l}}$ and $E_{l} \in \mathbb{R}^{M \times F_{l}}$.
The latent node and edges are given by
\begin{equation}
    \begin{aligned}
        V_{l} &= \varphi_v(V_{i}), \\
        E_{l} &= \varphi_e(E_{i}),
    \end{aligned}
\end{equation}
where $\varphi_v$ and $\varphi_e$ are multi-layer perceptrons (MLPs). 

\paragraph{Processor} The processor consists of $K$ interaction networks (IN) \cite{Battaglia2016_interaction} such that the output attributes of one network are summed to its input attributes to produce a new latent graph. The new latent graph is given as input to the next network to propagate information through the graph. The output of the processor is the final latent graph $G_{K}=(V_K, E_K)$. Each IN is defined as
\begin{equation}
    \begin{aligned}
        \mathbf{e}'_{ij} &= \phi_e(\mathbf{e}_{ij}, \mathbf{v}_{i}, \mathbf{v}_{j}),   &  \mathbf{\overline{e}}'_{j} = \sum_{ \mathbf{e}_{ij} \in E_{j} } \mathbf{e}_{ij}, \\
        \mathbf{v}'_{i} &= \phi_v(\mathbf{v}_i, \mathbf{\overline{e}}'_{i}), & 
    \end{aligned}
\end{equation}
where $E_j = \{ \mathbf{e}_{ij} \mid i = 1,\cdots,N \}$ is the set of edges whose receiver node is the node with index $j$. The functions $\phi_e$ and $\phi_v$ are MLPs such that each IN has the same architecture but different weights. 

\paragraph{Decoder} The decoder produces the output for each node using the nodes of the graph produced by the final IN of the processor by
\begin{equation}
    \mathbf{\ddot{x}}_i = \varphi_a(\mathbf{v}_i), \quad \mathbf{v}_i \in V_K,
\end{equation}
where $\mathbf{\ddot{x}}_i \in A$ is the acceleration of the particle and $\varphi_a$ is an MLP.

The predicted accelerations are used to update the next state and then compute the trajectory. The next state is computed using the semi-implicit Euler integration via
\begin{equation}
\label{eqn:semi-implicit-euler}
    \begin{aligned}
        \mathbf{\dot{x}}_{i,t} &= \mathbf{\dot{x}}_{i,t-1} + \Delta t \, \mathbf{\ddot{x}}_{i,t-1}, \\
        \mathbf{x}_{i,t} &= \mathbf{x}_{i,t-1} + \Delta t \, \mathbf{\dot{x}}_{i,t}.
    \end{aligned}
\end{equation}

\subsection{Trajectory planning for granular manipulation}
\label{sec:traj_opt}
Our objective is to control the rigid-body particles $P_r$ by following a trajectory $\mathcal{T}$ that leads the granular particles to settle into a desired goal position $P_{g,G}$. 
We consider a planning horizon $H$, where the granular particles start from an initial position $P_{g,0}$, located inside the rigid-body.
The rigid-body can be controlled through two set of actions: rotation and translation $\mathbf{u}_t=(\mathbf{R}_t, \mathbf{x}_t)$.
The optimal trajectory is defined by $Q$ via-points $\mathcal{T}^*=(\mathbf{u}_0, \dots, \mathbf{u}_Q)$. The trajectory is then interpolated to get the actions in the planning horizon $H$ that minimise the distance between the end position of the particles and the desired goal position:
\begin{equation}
    \mathbf{u}_0, \dots, \mathbf{u}_Q = \argmin d(P_{g,G},P_{g,H}),
\end{equation}
where $d(\cdot, \cdot)$ is the distance between the desired and predicted distributions of the particles.
We choose this approach as it is unnecessary to transport each particle to the exact position of that same particle in the target distribution, as long as the distributions match.
In contrast to existing approaches in particle manipulation that use the Chamfer distance, which measures the distance of every particle to its nearest neighbour \cite{Li2019_learning_particle_dynamics, ChenS2021_ab_particle_manipulation}, we propose to use the Wasserstein distance, which can be understood as the minimum cost of moving the particles from one distribution to the other.
We expect this to express better the semantics of material transport.
Thus, the trajectory planning is formulated as a discrete optimal transport (OT) problem, where we minimise the quadratic Wasserstein distance: 
\begin{equation}
    d(P_{g,G}, P_{g,H}) = \left(
    \sum_{i=0}^N \lVert X_{i,G} - X_{i,H} \rVert^2
    \right)^{1/2},
\end{equation}
where $X_{i,G} \in P_{g,G}$ and $X_{i,H} \in P_{g,H}$ are the samples of each empirical distribution.
One of the issues when computing the Wasserstein distance for a large number of particles is the computational cost.
In order to utilise the Wasserstein distance with large datasets we use the Sinkhorn loss $S_\varepsilon(P_{g,G},, P_{g,H})$, which is an approximation of OT and reduces the complexity as presented by \cite{feydy_2019_otsinkhorn}.

To solve the optimisation problem, we adopt a population-based black-box optimiser, the Covariance Matrix Adaptation Evolution Strategy (CMA-ES) \cite{hansen_2001_cmaes}, and treat the trajectory planning as a constrained optimisation problem.
As constraints, we use
the manipulator end-effector limits and the velocity limits used for training the model, similar to \cite{fares_2015_ga_traj_planning}.
The constraints guide the optimisation to consider trajectories within the training data of the model. 
The optimisation problem is then

\begin{equation}
\label{eqn:cma_es_argmin}
\begin{aligned}
    \min_{\mathbf{u}_0, \dots, \mathbf{u}_Q }
    \quad  &J(P_{g,G}, P_{g,H}, \mathcal{T})\\
    \text{subject to} \quad & |\mathbf{u}_q -  \mathbf{u}_{q-1}| \leq \Delta \mathbf{u}_{max},\\
    & \mathbf{u}_{min} \leq \mathbf{u}_q \leq \mathbf{u}_{max},
\end{aligned}
\end{equation}
for all $q$ = $0, \dots, Q$, where $J(\cdot)$ is the cost function and $\mathbf{u}_{min}$, $\mathbf{u}_{max}$ and $\Delta \mathbf{u}_{max}$ are the boundaries and velocity limits respectively.
The cost function, in addition to minimising the distance between experimental distributions, includes a term that penalises the acceleration of the rigid-body to reduces abrupt motions, such that
\begin{equation}
\label{eqn:cost_function}
\begin{aligned}
    J(P_{g,G}, P_{g,H}, \mathcal{T}) &=
    \alpha \, S_\varepsilon( P_{g,G}, P_{g,H})\\
    &+ \beta \sum_{t=2}^H \lVert \mathbf{u}_t - 2\mathbf{u}_{t-1} + \mathbf{u}_{t-2} \rVert^2
\end{aligned}
\end{equation}
where $\alpha$ and $\beta$ are regularisation constants.

The optimal trajectory is found by following Algorithm~\ref{alg:cma-es}. The algorithm takes as input an initial trajectory $\mathcal{T}$ as well as the initial and target particle configurations $P_{g,0}$ and $P_{g,G}$. Then, the next position of the rigid-body particles $P_{r,t}$ is computed after applying the actions $\mathbf{u}_t$. From this, the control inputs $\mathbf{c}_t$ are computed as the difference of the rigid-body position $P_{r,t}$ and $P_{r,t-1}$. Next, the graph $G_{i,t}$ is created as explained in Section~\ref{sec:graph_represent}. The next position of granular particles $P_{g,t+1}$ is updated on line 5 by following the semi-implicit Euler formulation \eqref{eqn:semi-implicit-euler}.
Once the rollout has finished, the cost $J(\cdot)$ is computed. We perform the optimisation $T$ times, where CMA-ES generates a new population until the maximum number of iterations is reached.

\noindent\begin{minipage}[t]{0.48\textwidth}
\begin{center}
\vspace{-0.15cm}
\begin{algorithm}[H]
\caption{Trajectory planning}
\label{alg:cma-es}
\begin{algorithmic}[1]
\Require $P_{g,0}$, $P_{g,G}$, $\mathcal{T}$
\Ensure $\mathcal{T}^*$
\For{$i=0 \rightarrow T$}
    \For{$t=0 \rightarrow H$}
        \State $\mathbf{u}^*_t \gets \mathcal{T}_t$ 
        \State $G_{i, t} \gets$create\_graph$(P_{g,t}, \mathbf{u}^*_t)$  \Comment{\hspace{-0.1cm}Section~\ref{sec:graph_represent}}
        \State $P_{g,t+1} \gets$GNN$(G_{i, t})$
    \EndFor
    \State $\mathcal{T} = \argmin J(P_{g,G}, P_{g,H}, \mathcal{T})$  \Comment{Using \eqref{eqn:cma_es_argmin}}
\EndFor
\State $\mathcal{T}^* \gets \mathcal{T}$
\end{algorithmic}
\end{algorithm}
\vspace{0.15cm}
\end{center}
\end{minipage}

\section{Experimental Results}

In this section, we first provide information about the physics engine simulation set-up as well as the GNN training and the trajectory planning routine.
Then, we perform an ablation study of the graph attributes and GNN parameters, where we evaluate the accuracy of the models predictions.
After that, we assess the trajectory planning routine accuracy for pouring the granular material into different goal positions.
The goals of the experiments are twofold:
1) to assess the required parameters for predicting accurately the interaction of granular materials with rigid-bodies,
2) to evaluate whether the proposed method can pour granular material into a desired shape, both in simulation and in a physical system.

\subsection{Simulation set-up}
\label{sec:sim-set-up}
We use Taichi-MPM \cite{Hu2018_mls_mpm} to generate training data for the GNN model.
We create a simulated scene where a cup pours granular material into a container, which acts as boundaries of the scene.
The container is a $10 \times 20 \times 20$ cm box with an open top allowing the cup and granular material to be located above the container.
The cup is modelled as a 3D mesh of $\varnothing 7 \times 10$ cm, which has been populated and replaced by rigid-body particles $P_r$.
The cup is controlled via translation and rotation actions.
The granular material is modelled as sand particles which are initialised inside the cup.
The simulation consists of $H=300$ time-steps, where the number of particles is 1945, out of which $70\%$ represent the granular material and the rest the rigid-body particles.

\begin{figure*}
    \centering
    \vspace{0.05cm}
    \includegraphics[width=2\columnwidth]{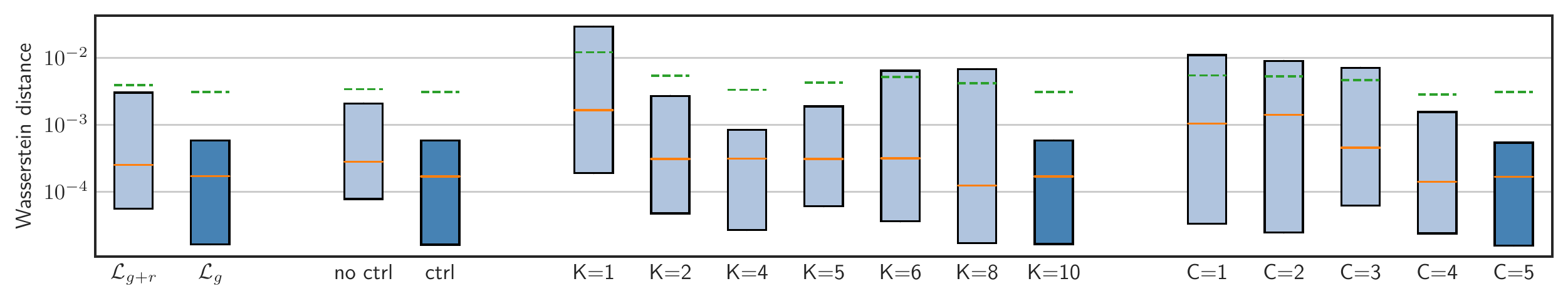}
    \caption{Boxplot of ablation study using forecast Wasserstein distance. Boxes represent values between 1st and 3rd quartiles, orange horizontal lines represent medians and dotted green lines represent means.
    Models marked with darker blue represent the GNN model with the proposed graph attributes and architecture.}
    \label{fig:ablation_study}
\end{figure*}

\subsection{GNN training details}
\label{sec:exp-gnn-details}
The training dataset consists of 20 simulations where the cup is translated along the $Y$-axis and rotated along the $X$-axis.
The translation and rotation axes are constrained so that the granular material is always poured inside the container.
We provide details about the different trajectories used for generating the training data in the Appendix. In addition, we have included a simulation where the cup only translates and rotates with small Gaussian noise and a simulation of the granular material falling without a cup in the scene.

The connectivity radius used to construct the graph is $R=0.015$.
Before constructing the graph we add a random-walk noise with standard deviation $\sigma = 3 \cdot 10^{-4}$ to the particle positions to improve the acceleration prediction as shown by \cite{Sanchez-Gonzalez2020_learning_to_simulate}. The velocities and accelerations are normalised with mean and standard deviation of the dataset. The distance to boundaries $\mathbf{b}$ in node attributes and displacement $\mathbf{s}_{ij}$ in edge attributes are normalised using the connectivity radius $R$, and $\mathbf{b}$ is clipped to range $[-1,1]$ as in \cite{Sanchez-Gonzalez2020_learning_to_simulate}.
The model is trained for $2000$ epochs using the Adam optimiser \cite{Diederik_Adam_2015}. For the first $500$ epochs the learning rate is constant $lr=10^{-4}$, after which we switch to an exponential learning rate with $\gamma = 0.997$.
The training is done using minibatches of size $2$. 
For each MLP in our model, we use two hidden layers where each hidden layer has a size of $128$. Each MLP in the encoder and processor are followed by a Layer Normalization \cite{Ba2016_layer_norm}. 
We have experimented using two different loss functions, the number of message-passing steps, including and excluding control inputs $\mathbf{c}_t$, as well as the length of velocity history $\{\mathbf{\dot{x}}_{t-C}, \dots, \mathbf{\dot{x}}_{t-1} \}$ in the node attributes.
The first loss function evaluated, $\mathcal{L}_g$, is the L1-loss that considers only accelerations of granular material particles. The second one, $\mathcal{L}_{g+r}$, is the L1-loss of both granular material and rigid-body particle acceleration. Both losses are averaged over all particles in the minibatch. 
All GNN models were trained until convergence, which takes 12 days for the model with parameters $K=10$ and $C=5$ on a computer with a NVIDIA V100 GPU.

\subsection{Trajectory planning details}
\label{sec:details-trajectory-optim}
We use four target distributions $P_{g,G}$ of the granular material for evaluating the optimal trajectory determined using Algorithm~\ref{alg:cma-es}.
The target distributions are generated from simulations that are not included in the training data of the GNN model.
The trajectory $\mathcal{T}$ has $Q=6$ via-points and it is interpolated using a piecewise cubic hermite interpolating polynomial to the rollout horizon length used in the GNN training and test models.
For the CMA-ES optimiser we used an initial variance of $\sigma_\mathcal{T}=1.5$ and a population size of 20. In order to speed up the optimisation process, we scaled the trajectory rotation by $\pi$ and the translation by $0.11$ so that both variables are within the same range. The algorithm is run for $150$ iterations with five different random seeds, where each random seed took 10 hours on a computer with a V100.
The constraints were defined as $\mathbf{u}_{min, max}=\{\pm2.8973, \pm0.1\}$ and $\Delta\mathbf{u}_{max}=\{2.1973, 0.02\}$, which matches the maximum values of the training data distribution.
We used $\alpha=1000$ and $\beta=0.001$ as values for the regularisation terms of the cost function \eqref{eqn:cost_function}. The $\alpha$ parameter is scaled with a high value due to the order of magnitude of the Sinkhorn loss, whereas $\beta$ is reduced to keep it as a second minimisation objective. The values for these parameters were determined experimentally. We used one of the sinusoidal trajectories used for training (see Appendix) as initial trajectory $\mathcal{T}$, where the final material distribution does not match the test cases target.

\subsection{GNN ablation study}
\label{sec:gnn-ablation}
We perform an ablation study of the aforementioned parameters.
To evaluate the accuracy of the predictions, we use the Wasserstein distance between the predicted granular material particles and the ground-truth simulation. 
The initial Wasserstein distance for simulations when the particles are relatively still in the cup and the forecast error has not yet accumulated is in the order of $10^{-5}$. 
In our experimental set-up we can consider that a Wasserstein distance of $10^{-2}$ for the entire rollout presents distant distributions, whereas a value in the order of $10^{-4}$ accurately matches the rollout particle distributions.
The tests are performed by starting from the same initial state onward to create a rollout prediction, where the rollout length is the same as the simulation.
The test-set includes 8 simulations that are in the model training domain.
The results are summarised in Fig.~\ref{fig:ablation_study}, where
the box plot excludes minimum and maximum values, since they are similar in each tested model. In Fig.~\ref{fig:ablation_study} the best performing model is highlighted in darker blue, which was used to rollout the dynamics in the trajectory planning.

\paragraph{Training loss}
We experimented on two loss functions $\mathcal{L}_{g}$ and $\mathcal{L}_{g+r}$.
As shown in Fig.~\ref{fig:ablation_study}, the loss $\mathcal{L}_{g}$ presents a slightly lower median and mean.
The most noticeable difference between the two models is their quartiles, which are lower for the loss $\mathcal{L}_{g}$.
This provides evidence that including the acceleration of rigid-body particles in the loss is unnecessary as it affects adversely when learning granular particle behaviour. 

\paragraph{Control inputs}
We investigated whether including control inputs $\mathbf{c}_{t}$ as a node attribute improves the model accuracy.
Considering the quartiles of the models in Fig.~\ref{fig:ablation_study} we can observe lower loss for the model that includes control inputs.
This is to be expected since it gives additional information about the movement of the rigid-body, which improves the predictions of the granular material particles.

\paragraph{Message-passing steps}
We also experimented on the number of INs in the processor. \cite{Sanchez-Gonzalez2020_learning_to_simulate} suggests that using a higher number of message-passing steps provides better results. We selected $K=10$ as maximum and evaluated how models with smaller values perform. The range of values selected was $K \in \{ 1,2,4,5,6,8,10 \}$.
We can notice in Fig.~\ref{fig:ablation_study} that the highest $K$ value indeed does perform best. 
The message-passing steps affect on how far the particle information reaches in the graph and, thus, how many interactions affect the particles. 
With $K=4$ the performance is not significantly worse than $K=10$.
This suggests that $K=4$ message-passing steps is enough to capture the system behaviour. However, the third quartile of $K=6$ and $K=8$ are higher than for $K=4$, which suggests otherwise.

\paragraph{Node velocity history length}
Finally, we assessed how the number of previous velocities included in the node attribute affects the performance. The values we tested were $C \in \{ 1, 2, 3, 4, 5 \}$.
One would hypothesise that $C=3$ would be sufficient to capture the dynamics, as granular materials can be approximated via a second-order system.
However, the performance of $C=4$ and $C=5$ is significantly better than the other models. This suggests that knowing the three previous velocities is not enough to accurately predict rollout dynamics in our system. 

Overall, the results show that including the control input and the $\mathcal{L}_{g}$ loss function yield drastically lower values compared to the other variants, as the interquartile range improves by almost one order of magnitude.
The results for the number of message-passing steps showed that with $K=10$ the mean and interquartile range are lowest within the tested models. 
The node velocity history length $C=5$ provided slightly higher mean and median than $C=4$, but presented lowest interquartile range. For this reason we choose to use $C=5$, $K=10$, control inputs and $\mathcal{L}_{g}$ as loss function for the GNN model in the trajectory planning.
This model speeds up the computation of the rollout dynamics by 20 times the high-fidelity simulator.

\subsection{Trajectory planning results}
\label{sec:results-traj-optim}
\begin{figure}
    \centering
    \subfloat[Test Case 1]{\includegraphics[width=0.45\columnwidth]{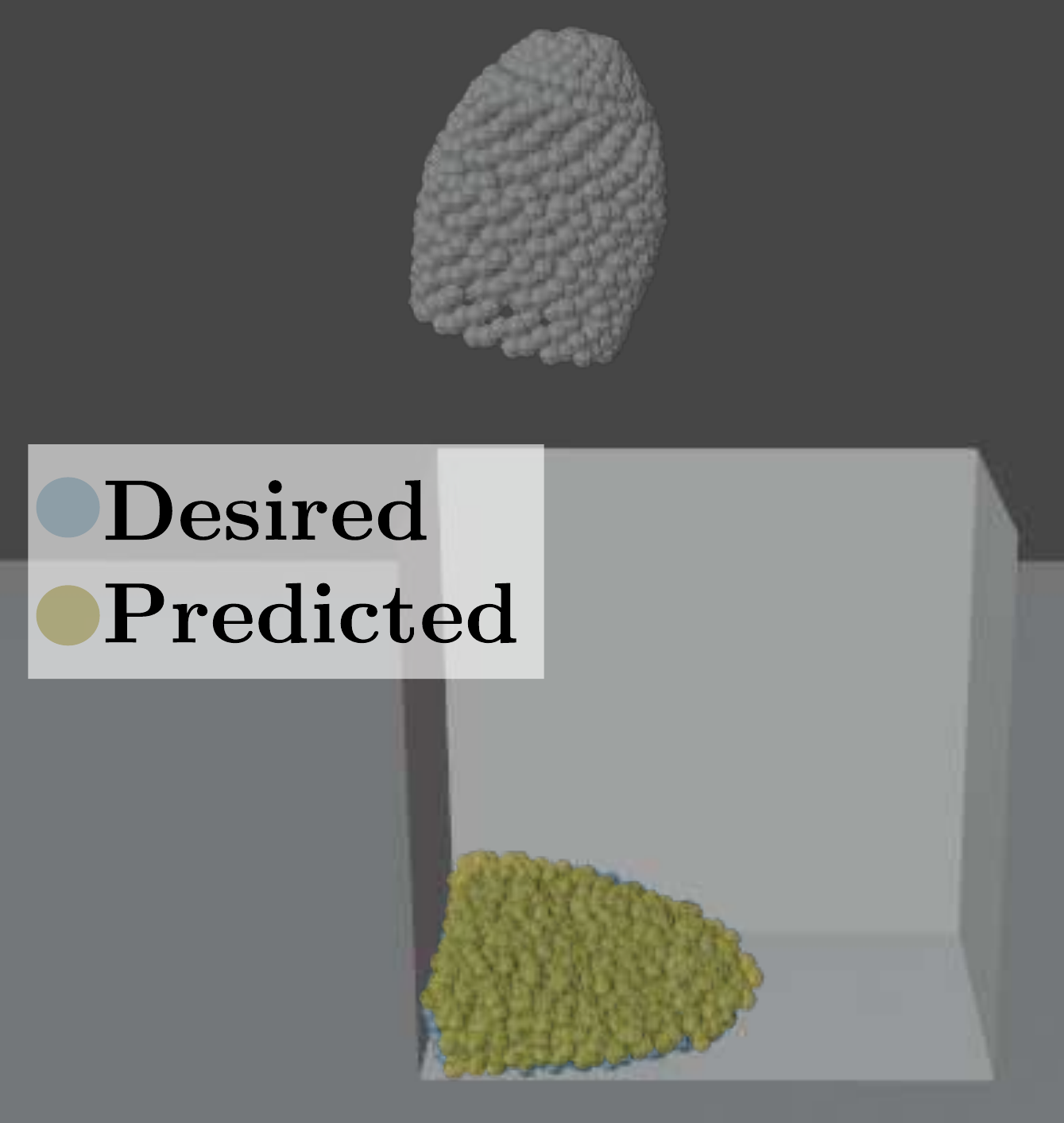}} \hspace{0.2cm}
    \subfloat[Test Case 2]{\includegraphics[width=0.45\columnwidth]{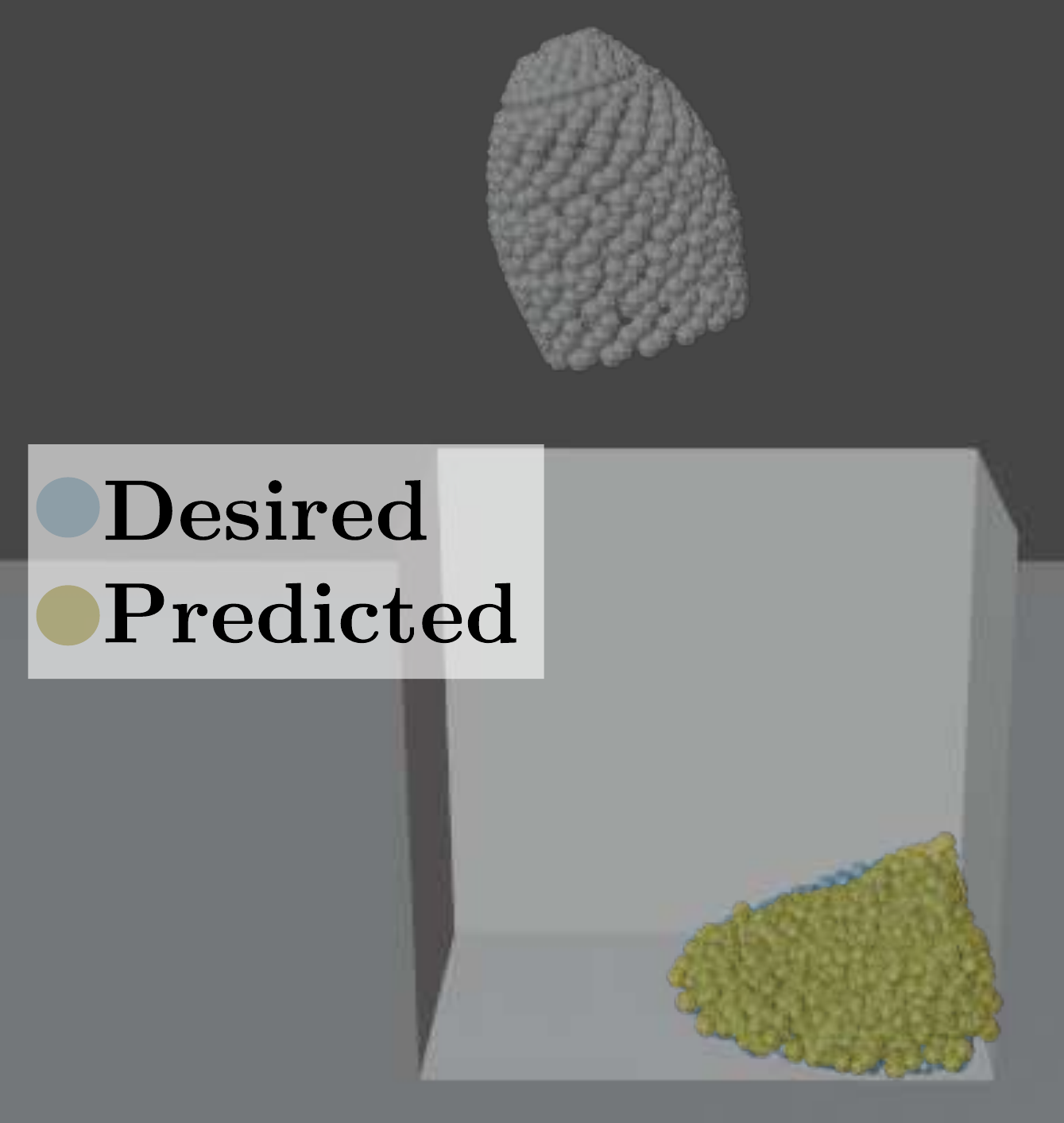}}\\
    \vspace{0.2cm}
    \subfloat[Test Case 3]{\includegraphics[width=0.45\columnwidth]{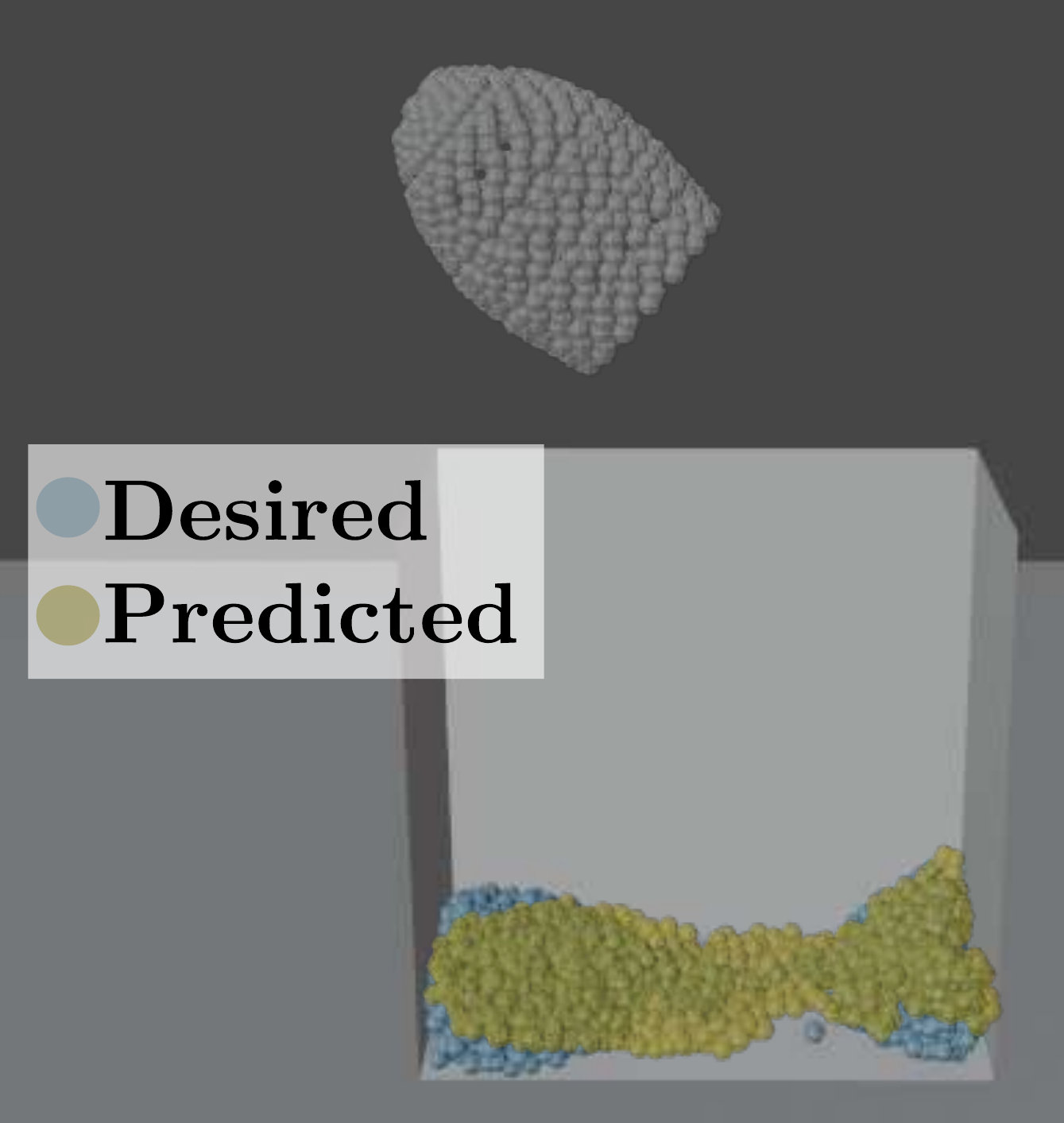}} \hspace{0.2cm}
    \subfloat[Test Case 4]{\includegraphics[width=0.45\columnwidth]{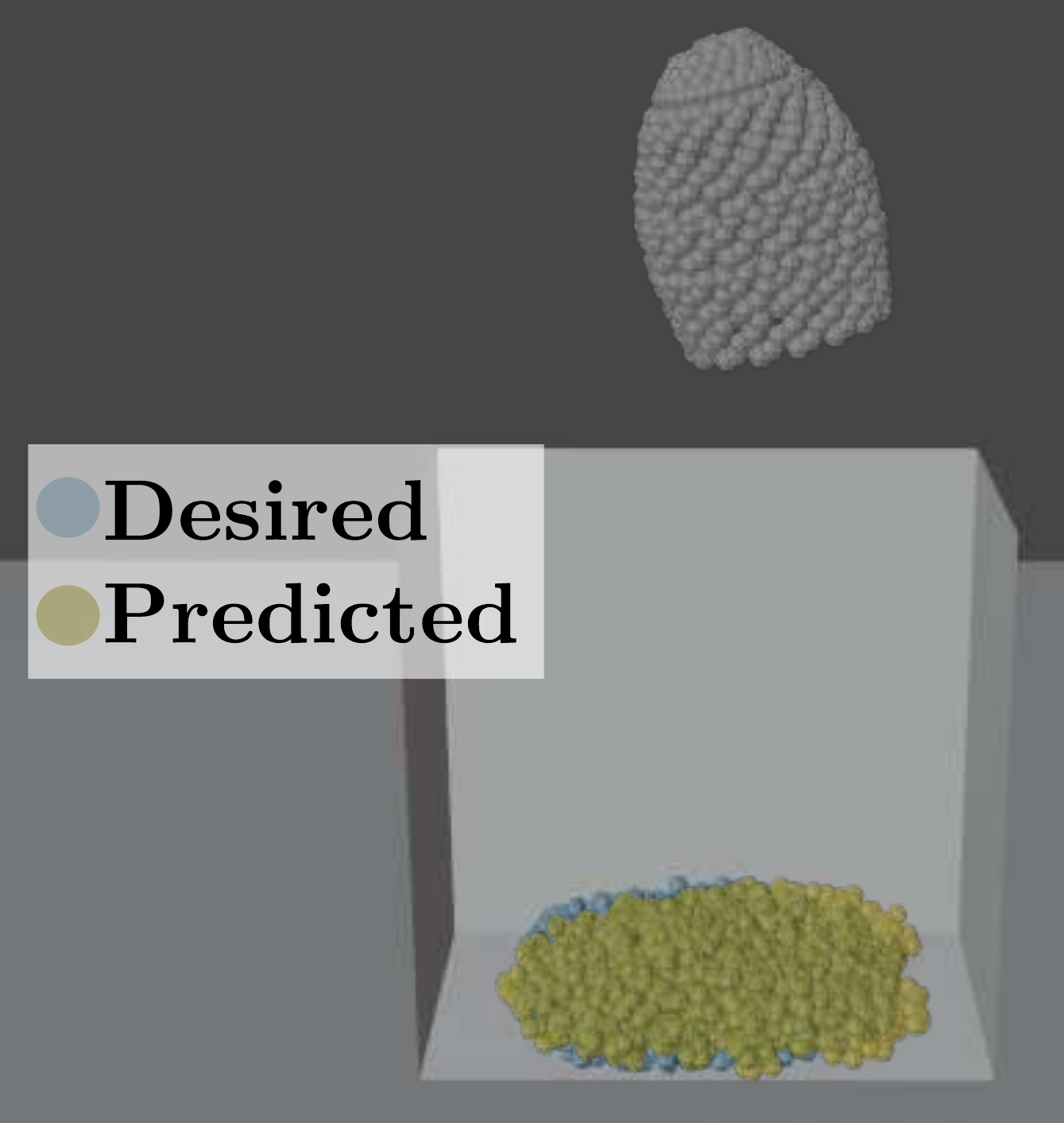}}
    \caption{\textbf{Qualitative results} for each test case showing the granular material final configuration (predicted) and goal configuration (desired) after following an optimal trajectory.}
    \label{fig:pouring_fig}
\end{figure}

The Sinkhorn loss for each of the four test cases is shown in Table~\ref{tab:traj-opt-results}.
The initial loss $S_\varepsilon(P_{g,G}, P_{g,0})$ measures the distance between the distribution of the particles in the initial configuration of the cup and its target.
We also provide the loss for the end distribution of the particles following the initial trajectory  $P_{g,H}$. This is used as a reference to highlight the improvement on matching the target distribution using the proposed algorithm.
We can notice that the order of magnitude improves by an order of two following the optimal trajectory $P_{g,H^*}$.
Our experiments showed that two particles distributions are well matched when the Wasserstein distance is in the order of $10^{-6}$, whereas distances larger than $10^{-4}$ are far from the target distribution.
The results indicate that our method is able to consistently reduce the distance between the end and target particle distributions by an order of magnitude.
The particles distribution after following the optimal trajectory as well as the desired distribution for each test case are shown in Figure~\ref{fig:pouring_fig}.
The final distribution for the test cases one, two and four nicely overlays the desired configuration of the material. On the other hand, the third test case Sinkhorn loss is reduced by an order of magnitude from the initial loss (see Table~\ref{tab:traj-opt-results}). However, the final configuration does not accurately match the target shape. We hypothesise that with a larger population of CMA-ES the Algorithm~\ref{alg:cma-es} would find a better solution.

\begin{table}
\renewcommand{\arraystretch}{1.8}
\vspace{0.15cm}
\caption{Sinkhorn loss between the target shape $P_{g,G}$, the initial material distribution $P_{g,0}$, the final distribution following the initial trajectory $P_{g,H}$, and the optimal trajectory $P_{g,H^*}$, for each test set over five different random seeds.}
\centering
\begin{tabular}{c c c c}
\toprule
\bfseries Test & $S_\varepsilon(P_{g,G}, P_{g,0})$ & $S_\varepsilon(P_{g,G}, P_{g,H})$  & $S_\varepsilon(P_{g,G}, P_{g,H^*})$ \\
\toprule
1 & $3.90\cdot 10^{-2}$ & $1.43\cdot 10^{-4}$ & $(2.09 \pm 0.80) \cdot 10^{-6}$\\
2 & $3.87\cdot 10^{-2}$ & $5.68\cdot 10^{-3}$ & $(1.16 \pm 0.36) \cdot 10^{-6}$\\
3 & $3.93\cdot 10^{-2}$ & $2.29\cdot 10^{-3}$ & $(1.88 \pm 0.53) \cdot 10^{-4}$\\ 
4 & $3.83\cdot 10^{-2}$ & $1.57\cdot 10^{-3}$ & $(2.69 \pm 0.78) \cdot 10^{-6}$\\
\bottomrule
\end{tabular}
\label{tab:traj-opt-results}
\vspace{0.2cm}
\end{table}

\begin{figure*}
    \centering
    \vspace{0.05cm}
    \subfloat[Test Case 1]{\includegraphics[width=0.50\columnwidth]{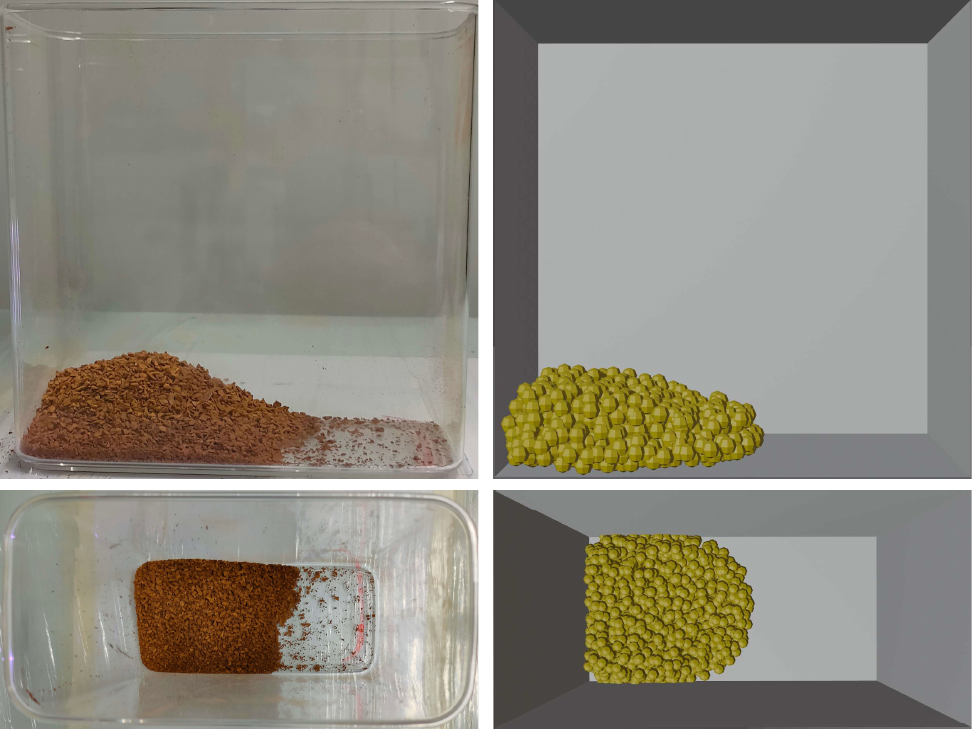}}
    \hspace{0.005cm}
    \subfloat[Test Case 2]{\includegraphics[width=0.50\columnwidth]{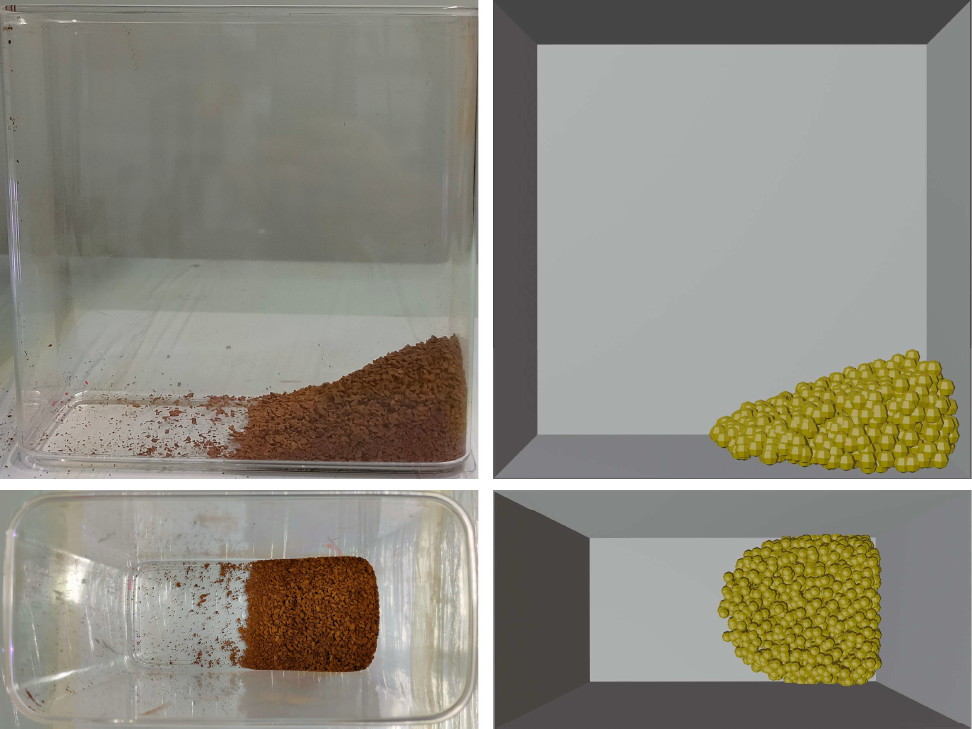}}
    \hspace{0.005cm}
    \subfloat[Test Case 3]{\includegraphics[width=0.50\columnwidth]{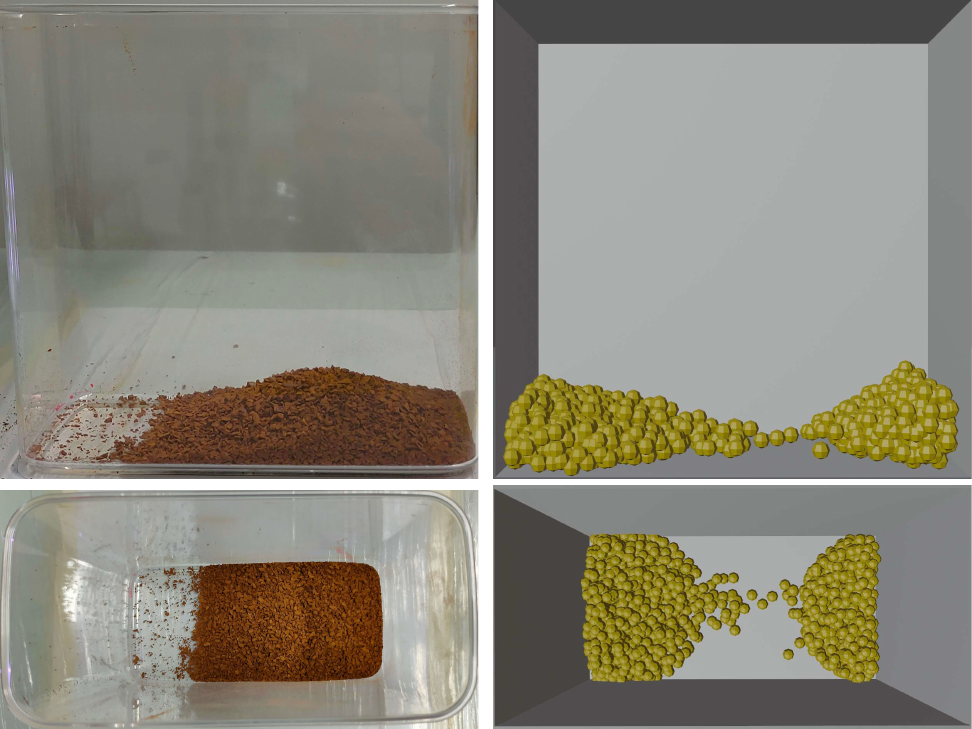}}
    \hspace{0.005cm}
    \subfloat[Test Case 4]{\includegraphics[width=0.50\columnwidth]{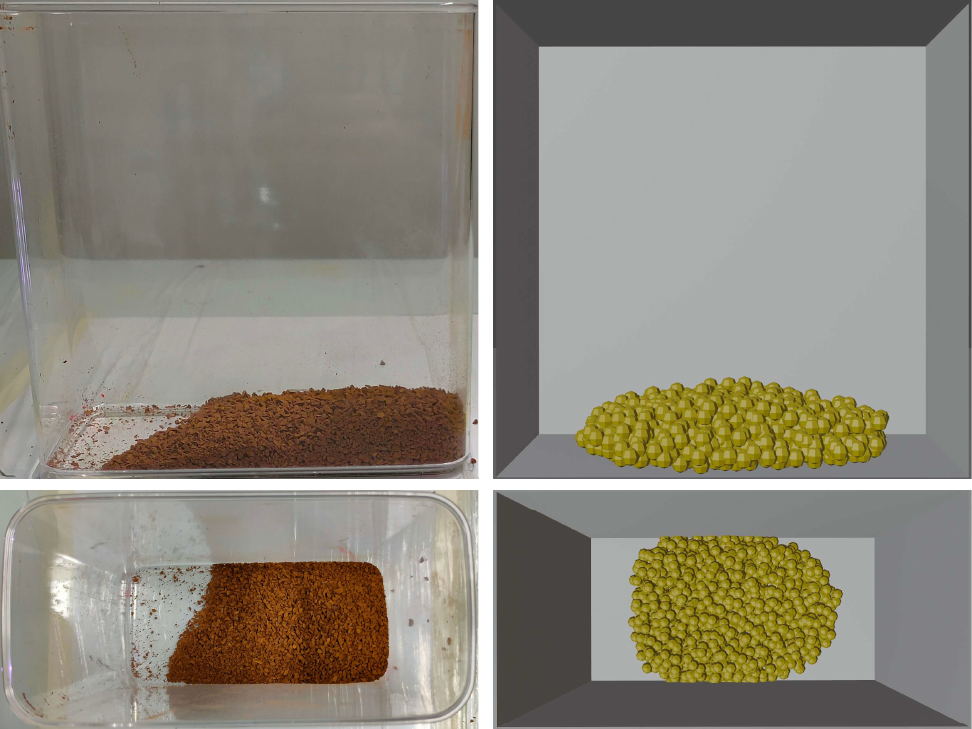}}
    \caption{\textbf{Qualitative results} of pouring ground coffee. The images show front and top views of the particles distribution after following the optimised trajectories on the real robot (\textbf{left}) and target distribution using the ground-truth  simulation (\textbf{right)}.}
    \label{fig:real_results}
\end{figure*}

We noticed that our GNN model was unable to generalise to rotation and angular velocities greater beyond those seen during training, where the model would predict the material leaking from the rigid-body (see supplementary material).
This limited the range of actions the trajectory planning algorithm could apply. Extensions of this work could focus on finding a model that generalises to any rigid-body action or training the model with a wider range of actions. 

\subsection{Granular manipulation in real-world}

We perform experiments on a real set-up where we use a Franka Emika Panda robot to manipulate the cup. We selected ground coffee as the granular material to pour into the container.
We executed the best four trajectories optimised in simulation using the real robot and qualitatively evaluated the end result of the ground coffee particles.
Prior to running the trajectories, we transform the reference frame of the rigid-body into the robot end-effector reference frame. The robot is controlled via a Cartesian controller \cite{khatib_1987_osc}, where each trajectory point is interpolated from the simulation frequency (100Hz) to the robot frequency (1KHz).

We show the qualitative results in Figure~\ref{fig:real_results}.
We can observe that the first and second test cases, (Figure~\ref{fig:real_results} (a), Figure~\ref{fig:real_results} (b)), match closely the desired distribution.
In contrast, the fourth test case, Figure~\ref{fig:real_results} (d), slightly differs from the target distribution.
Furthermore, the third test case, Figure~\ref{fig:real_results} (c), which required pouring in both left and right side of the container was the distribution furthest to the target.
In addition, in the supplementary material we provide the real-world performance of the predefined trajectories used for generating the target distributions. These show that the distributions generated in the simulator do not exactly match the real-world, with some spilling of the material for the first test case, and a slight mismatch for the third and fourth test cases.
This provides further evidence of the sim-to-real gap between the simulated and the real material.

As a summary, our results provide evidence that the proposed method is able to optimise the trajectory of the cup and pour the material in the desired configuration using the GNN rollout.
In addition, the results on the real-world highlight that the proposed method is able to effectively pour the material in some of the presented test cases without gathering data from the real-world to train the model.
However, some of the test cases were far from the results using the GNN rollout.
This is at least partially caused by imperfect simulation. Particles sometimes leak through the rigid body in the simulation, which is also illustrated in the supplementary material. We hypothesise that with a higher fidelity simulation that resembles better the granular material, such as the one proposed by \cite{matl_2020_infering_granular_properties}, the proposed model would be able to match better the real material and pour precisely the granular material. 

\section{Conclusion}

In this paper, we introduced a model-based approach for manipulating granular materials into a desired configuration by interacting with a rigid-object.
The proposed method uses a GNN to learn the interactions of the granular material from simulations, as well as the interactions of the material with the rigid-body that manipulates it.
We presented a trajectory planning routine that uses the GNN model to rollout the dynamics and optimises the trajectory by minimising the Wasserstein distance between the rollout result and the goal distribution of the poured material.
We provided a study of the GNN model architecture as well as the graph attributes that can accurately predict the interactions between the material and the manipulated rigid-body. 
Our results show that the planning routine is able to find the required motions to pour the material in the desired configuration.
We also demonstrated that the optimal trajectories used with the GNN rollouts are able to effectively pour the material in a real-world set-up.

One of the main limitations of using data from a simulator for planning is the simulation-to-real world gap. Naturally, using a high fidelity simulation is beneficial but the calibration of any simulation with a particular real world setting is difficult. Thus, it is an important avenue for future  to study how simulation can be combined with limited data from the real world in order to achieve complex granular manipulation tasks across a variety of materials and tasks.

\section*{Acknowledgement}
The authors would like to thank Fares J. Abu-Dakka and Gokhan Alcan of Aalto University for their help and support in this work. The authors would also like to acknowledge the computational resources provided by the Aalto Science-IT project.

\appendix[Additional Training Details]

We defined four sets of trajectories equations to generate the data for training the Graph Neural Network:
\begin{enumerate}
    \item The first trajectory rotates following a cosine function with zero-mean Gaussian noise. The rotation stops for several timesteps once the maximum rotation is reached and then rotates back to upright position. The translation is given by a zero-mean Gaussian noise which is added to the position at the previous timestep.
    \item The second trajectory follows sinusoidal to translate and rotate the cup. 
    \item The third trajectory is given by a linear translation and rotation that tilt the cup a single time in both directions with constant velocity.
    \item The last trajectory is given by a linear translation and rotation with Gaussian noise $\mathcal{N}(0, 10^{-3})$ and $\mathcal{N}(0,1)$ respectively. The trajectory is defined as $300$ time-steps where the first third is limited to translation, the second to rotation, and the last to going back to the origin position and orientation.
\end{enumerate}

All the trajectories have randomised direction, maximum rotation angles and frequency for the sine and cosine trajectories, and maximum velocities for the linear trajectory. The specific details as well as the open source code can be found at: \url{https://sites.google.com/view/granular-gnn-manipulation}.

\bibliography{Granular_materials,liquid_manipulation,GNN, OT}
\bibliographystyle{IEEEtran}

\end{document}